\documentclass[letterpaper, 10 pt, conference]{ieeeconf}  
     \usepackage{hyperref}

\IEEEoverridecommandlockouts                              
\usepackage{graphics} 
\usepackage{epsfig} 
\usepackage{mathptmx} 
\usepackage{times} 
\usepackage{amsmath} 
\usepackage{amssymb}  
\usepackage{animate}
\usepackage{booktabs}
\usepackage{xmpmulti}

\title{\LARGE \bf
Vision Based Dynamic Offside Line Marker for Soccer Games
}

\author{Karthik Muthuraman\\
University of Michigan\\
{\tt\small mkarthik@umich.edu}
\and
Pranav Joshi\\
University of Michigan\\
{\tt\small pranavsj@umich.edu}
\and
Suraj Kiran Raman\\
University of Michigan\\
{\tt\small surajkra@umich.edu}
}

\begin{document}

\maketitle
\thispagestyle{empty}
\pagestyle{empty}

\begin{abstract}
Offside detection in soccer has emerged as one of the most important decision with an average of 50 offside decisions every game. False detections and rash calls adversely affect game conditions and in many cases drastically change the outcome of the game. The human eye has finite precision and can only discern a limited amount of detail in a given instance. Current offside decisions are made manually by sideline referees and tend to remain controversial in many games. This calls for automated offside detection techniques in order to assist accurate refereeing. In this work, we have explicitly used computer vision and image processing techniques like Hough transform, color similarity (quantization), graph connected components, and vanishing point ideas to identify the probable offside regions.

\end{abstract}

\section{INTRODUCTION}

Offside is a specific rule defined by the association of football (soccer), which states that a player is in an offside position if any of their body part except the hands and arms is in the opponents’ half of the pitch and closer to the opponents’ goal line than both the ball and the second-last opponent. 
	Traditionally offsides are called by the human-sideline referees who keep moving along the sidelines continuously tracking the last defender. Not only is the work of the sideline referee taxing, but this technique is prone to human errors at multiple levels due to the limited precision of human eye. Even though there are definite rules explaining the foul, offside call decisions are determined at the split second. Offside calls remain highly subjective to human referees and remains as one of the most complicated rules.

\begin{figure}[h]
\centering
\includegraphics[width=0.45\textwidth]{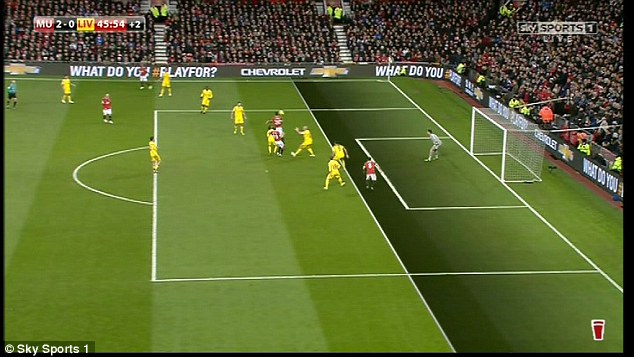}
\caption{An example of the offside rule/position}
\label{fig:offside}
\end{figure}

   Technology in sports has paved way for lot of improvements with respect to increased quality, accuracies and better decisions across multiple sports right from ball, player tracking to virtual simulations. Computer vision algorithms can be exploited efficiently for this problem of offside detection.
    
    Specifically in this problem of offside detection, techniques from Computer Vision can be used effectively to develop an end-end system of automatically determining offsides or can be used in assisting referees in making educated calls in almost real time. In this work, we present an assisstive technology to determine the offside line marker, in all frames of the real time video feed which could help in better decision making and accurate offside calls.
    More specifically, at any given frame of the video, we track all the players and determine which player is the `last' defender. Last defender here is defined as the player who is closest to the defending goal post, as seen along the vanishing line from that point. More details are mentioned later in this paper.

\section{Related Work}

Works related to offside detection specifically involve varying degrees of freedom with respect to the problem definition. Examples include assumptions on having a continuous feed of player locations on a 2D coordinate space, use of multiple camera views, use of static cameras, prior knowledge on the playing field region etc. 

Several works describe the process of identifying and tracking football players. Ming Xu et al.\cite{c1} use multiple static cameras set uniformly around the playing field to define player positions with each view and employ foreground identification and background subtraction techniques to effectively track across frames in a given video. Andres Galaviz et al. \cite{c2}employ image stitching techniques across multiple frames of a video and perform homography with a reference image to uniquely determine the player position in a given video input. A static camera is assumed.

Parth et al. \cite{c3}have developed an offside detection system by considering prior information on the team jerseys and field endpoints marked as inputs. The drawback of this method is the strict assumption of the still camera feed and the input of the field coordinates for the homographic transform and hence it is difficult to employ on a real life basis.

Jagjeet et al \cite{c4}have developed an offside detection system on still images. They have employed MobileNet, a pre-trained convolutional neural network for detecting players. This method is not robust across videos and with the use of Deep Learning models, they need high computational resources for efficient computation. 

An ideal offside tracker should be invariant to camera translations, stadium configurations, illuminations and player configurations and also have good real time performance . In this work, we present an offside line marker for every frame of a given video by employing player detection and tracking algorithms with image processing techniques capable of real-time implementations. No knowledge of the camera orientations nor any pre-trained models is assumed.We attack this problem solely based on visual information in the given frame/video. 

\section{Approach}

In this work, we have used FIFA 18 game-play videos which provide a fair approximation in simulating real-life soccer scenarios. The main challenges in this problem is the varying camera motion (translation), player kit configurations and stadium configurations. It is highly important to adaptively determine the play-areas, players and offside region for every given frame in the input feed. The only assumption in this setting is the presence of a camera which translated across different regions of the field with a constant zoom. 

\begin{figure}
\centering
\includegraphics[width=0.5\textwidth]{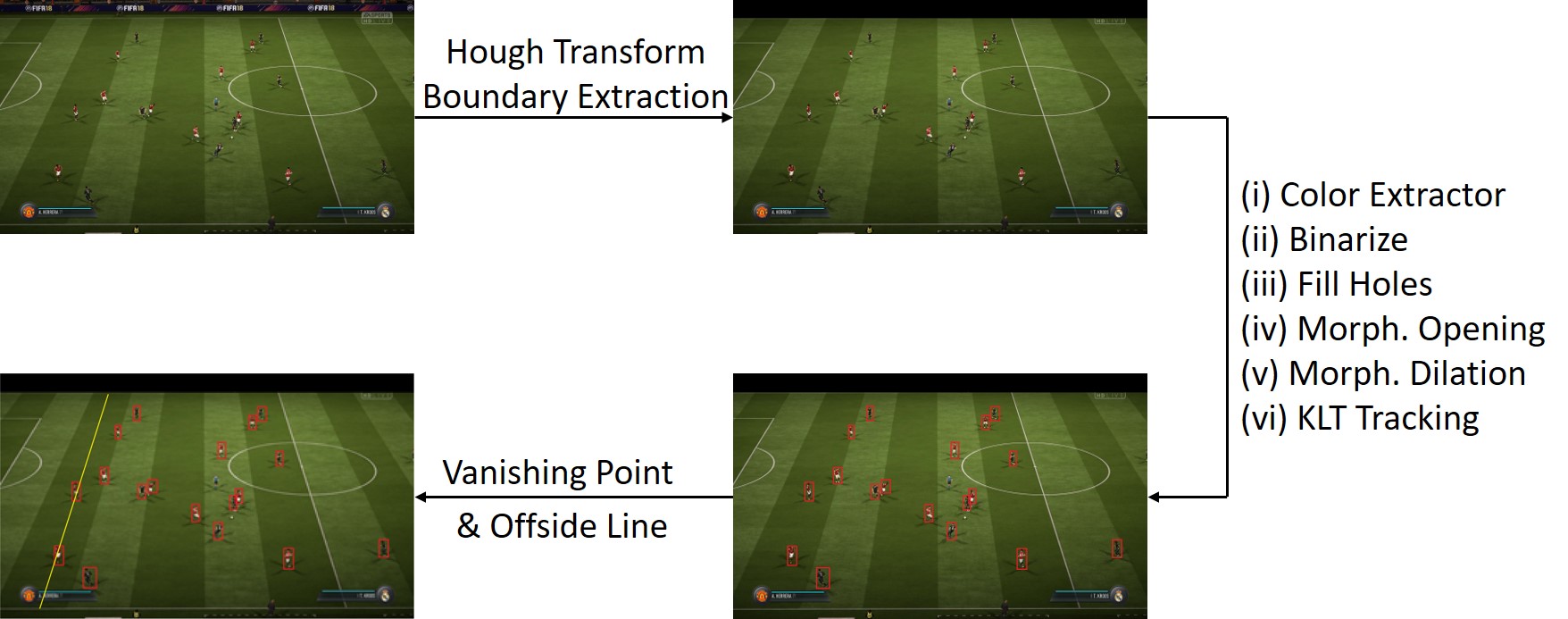}
\caption{Flow chart of our proposed method }
\label{fig:flow}
\end{figure}

\subsection{Determining the Play - Area}
In order to effectively track and identify players, it is important to segment out the field-area exactly. To be specific, we need to determine the top and side boundaries of an image to segment out the play-area. We exploited the presence of the line-boundary (white line) after identifying it using the Hough transform by carefully selecting only horizontal lines. The regions below this detected line is only considered as the field area. 

For the side boundary, we used the background color information (green) and a color similarity extractor to identify regions with similar color profiles. With image opening and filling operations, we extract only the significant connected components. The largest connected component corresponds to the green field region and hence we remove all other portions effectively removing the audience/ sign boards and identifying the play-area clearly. An example of this can be seen later in figure \ref{fig:side}.

\subsection{Identifying Players and Team Information}
With the team kit information and the processed field area from the above step, we use the color information of the team jerseys, and extract only those blobs which satisfy the color properties. We then binarize the image to form a mask of just the required colored pixels. In order to make the blobs continuous and overcome the effects of special designs on the player kit (kit numbers, sponsor logos etc), we perform an image filling operation wherein we fill holes to obtain a clear localization of the players of a given team. With several experiments we observed that players' other kit add-ons (shoes) sometimes form small components effectively qualifying as false positives in this detection problem. Thus we employed an image opening morphological operation to remove small insignificant components which is followed by image dilation to increase the strength of player detection. The final obtained image is a mask consisting of the detected players. In order to detect each player separately from the mask, we perform graph based connected component labeling wherein components that are connected are identified as foreground and hence bounding boxes for each player can be determined.

\subsection{Tracking the Players}
The main objective is to make real-time offside calls. Hence it is very important to have less complex models that can be implemented real-time. In order to improve the effectiveness, we include a tracking algorithm wherein we detect the players only once every second and we track the players for all the frames between successive detections. In a 30 fps video, we essentially detect players once and track them for the next 29 frames using the Kanade-Lucas-Tomasi tracker. The KLT algorithm works specifically well in our case because we are interested in tracking a small local region (bounding box) of a player in successive frames which correspond to minimal motion. The KLT algorithm was proposed mainly for solving an image registration problem wherein salient features in successive frames of a given region are identified and their motion in estimated in order to track the selected object. Good features are located by examining the minimum eigenvalue of each 2 by 2 gradient matrix, and features are tracked using a Newton-Raphson method of minimizing the difference between the two windows. In our work, we used the Harris features wherein we feed iterate through all bounding boxes of the players, determine Harris corner features in each local bounding box region. For each Harris corner, the motion (translation or affine) between consecutive frames is computed and they are linked in successive frames to get a track for each Harris point. This way we can optimize our algorithm and speed up the process by switching between detection and tracking.

\subsection{Drawing the offside line}
Once we have determined the player bounding boxes for each frame (either via direct detection or tracking), we now have to determine the offside line. To detect this, we need to first detect which defending player is closest to the goal along a vanishing line. Simply considering the `x' coordinate of the players will not work as the camera angle is not directly overhead. We make use of existing lines on the image and a novel method to determine the vanishing point. We extend each line and compute the pairwise intersections and average out to get the vanishing point. Once we have the vanishing point, we extend a line from the vanishing point to each defender and then select that line which has the lowest x intercept (from the left) on the bottom of the image. This way we get an accurate line. The figures below illustrate this point in detail.

\begin{figure}
\centering
\includegraphics[width=0.35\textwidth]{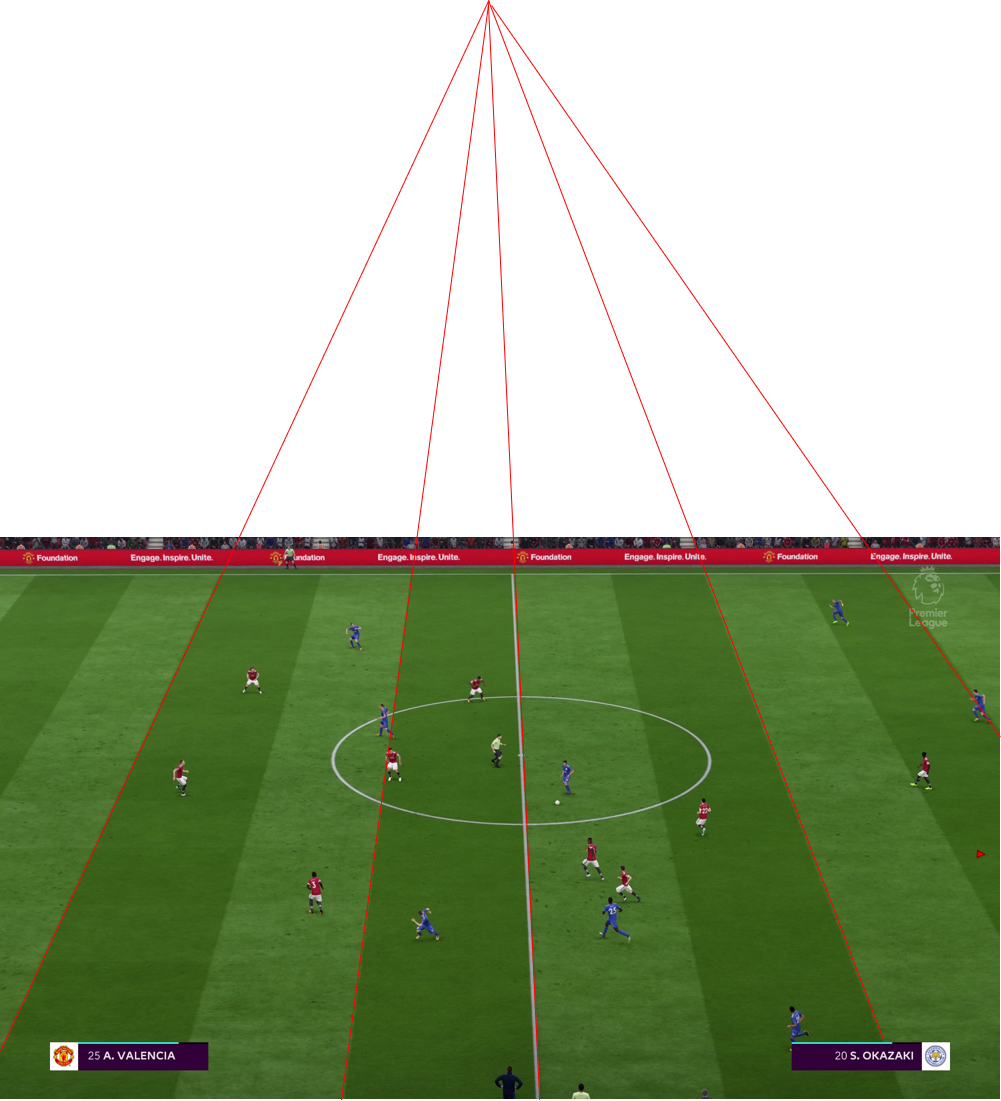}
\caption{Using lines on the field to determine vanishing point}
\label{fig:vp1}
\end{figure}

\begin{figure}
\centering
\includegraphics[width=0.35\textwidth]{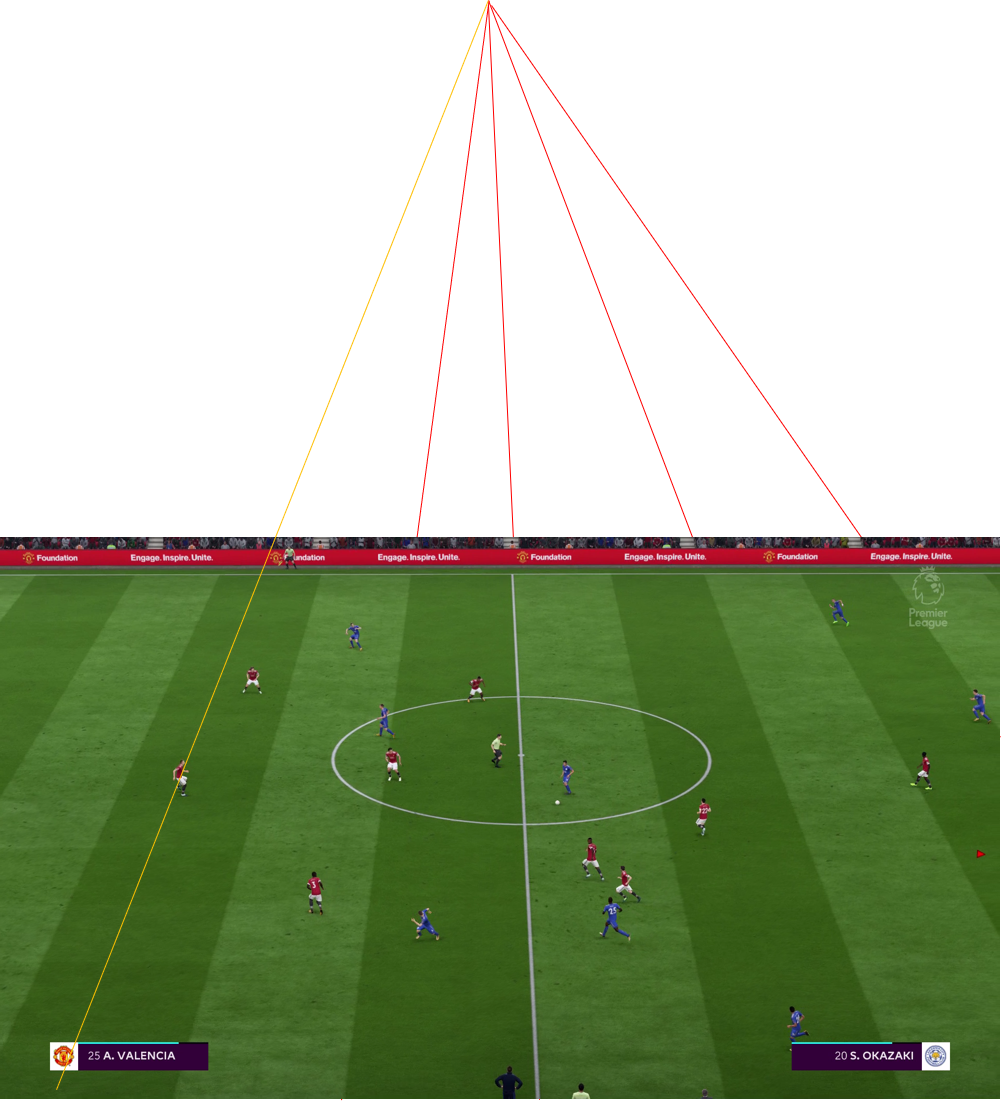}
\caption{Using the vanishing point to determine the offside line.}
\label{fig:vp2}
\end{figure}

As shown in figure \ref{fig:vp2}, the yellow line is the final offside line. It is the line joining the last player with the vanishing point. Note here that the last player is actually determined by joining the vanishing point to every player and then comparing the point where this line intersects the bottom of the image (x intercept). 

\section{Experimental Results}
\begin{figure}[h]
\centering
\includegraphics[width=0.45\textwidth]{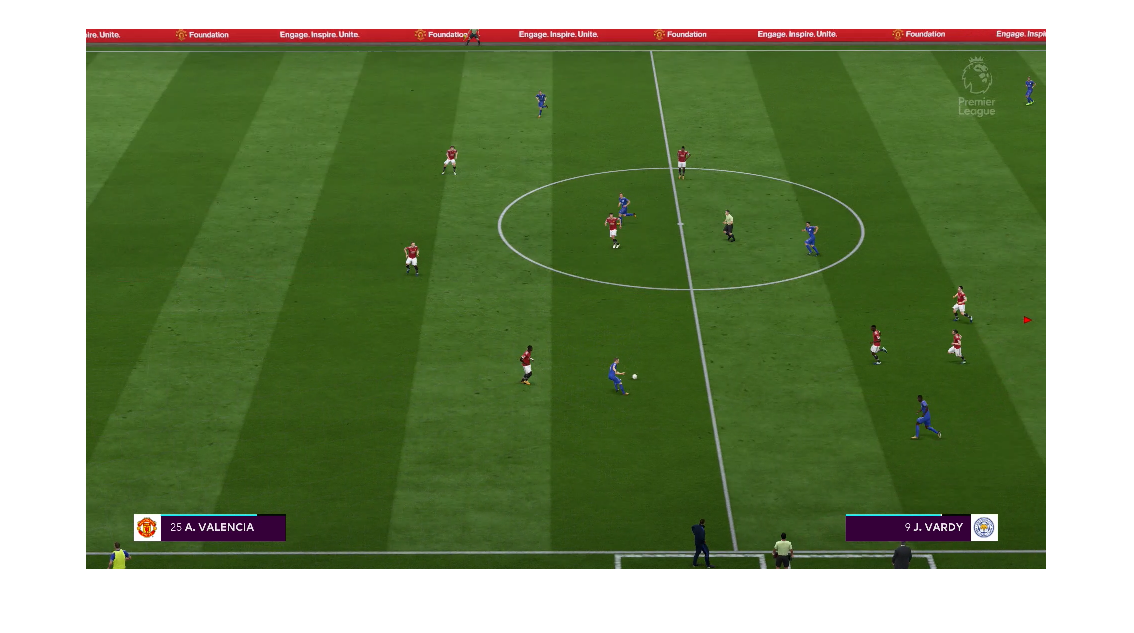}
\caption{Original Frame}
\label{fig:hough}
\end{figure}

\begin{figure}
\centering
\includegraphics[width=0.45\textwidth]{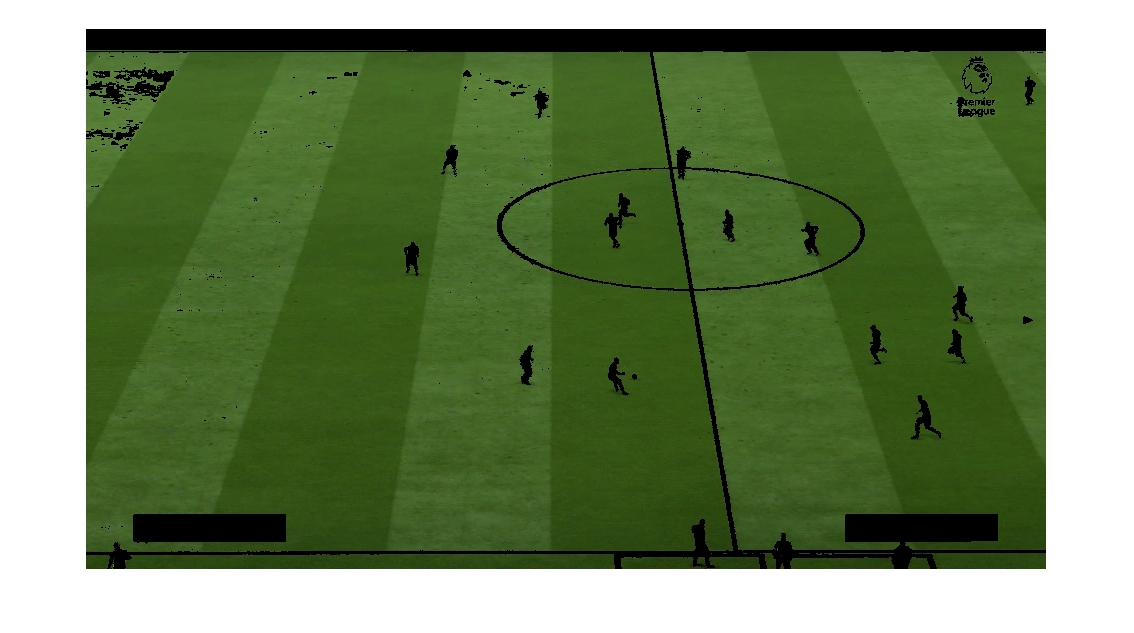}
\includegraphics[width=0.45\textwidth]{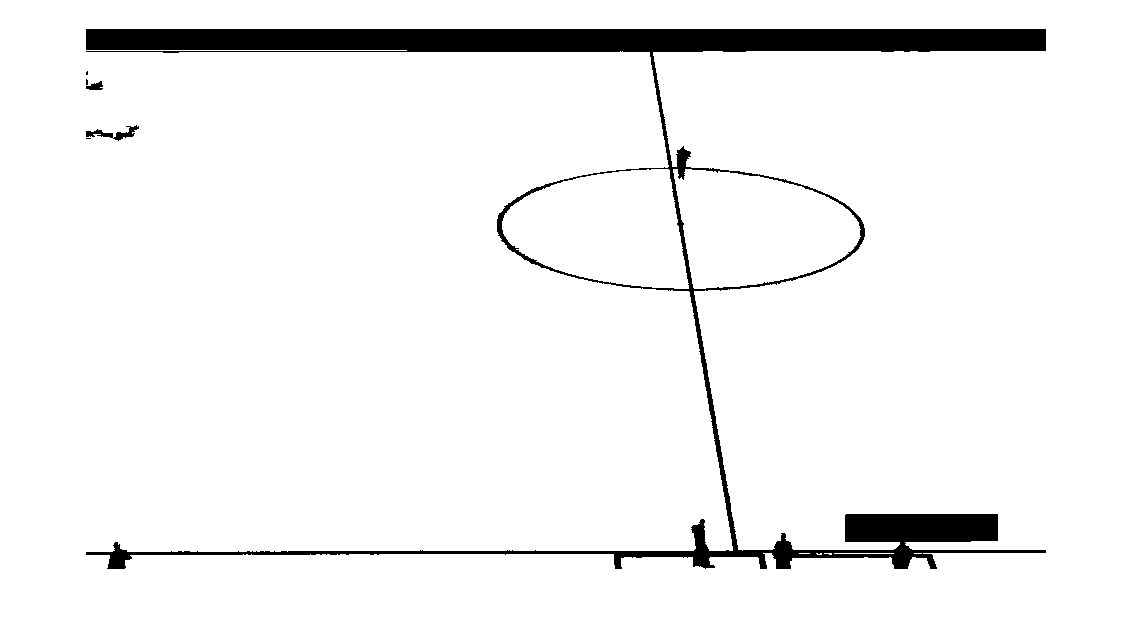}
\caption{(a) Hough transform used to remove top line and Green colours extracted to determine play area (b) Performing morphological operations on this play area to get a single largest connected component }
\label{fig:green}
\end{figure}

\begin{figure}
\centering
\includegraphics[width=0.45\textwidth]{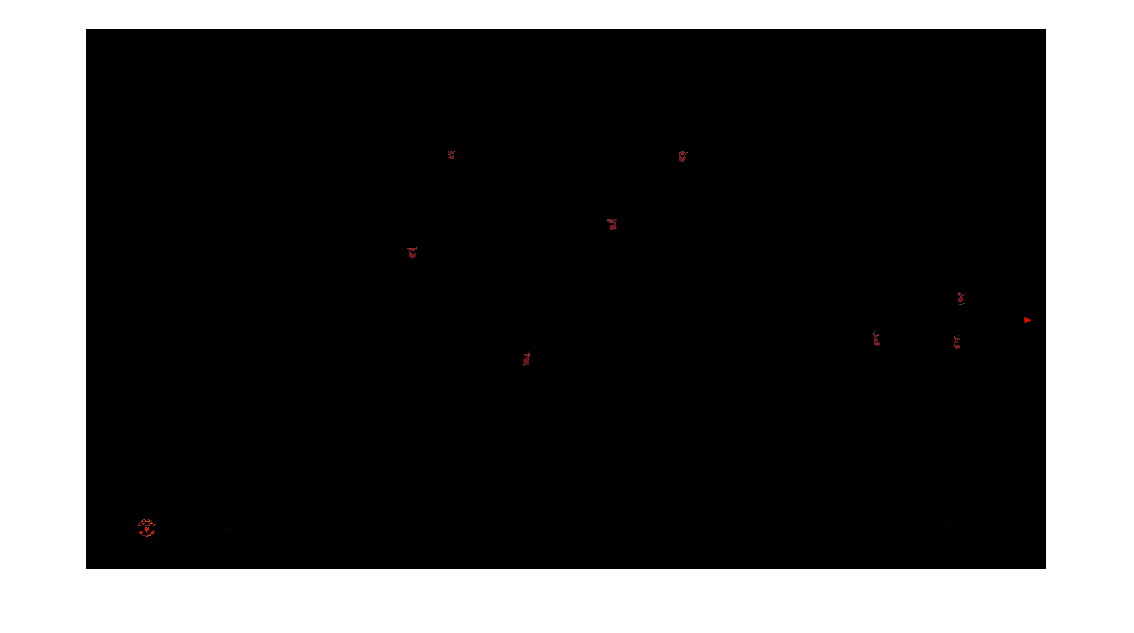}
\includegraphics[width=0.45\textwidth]{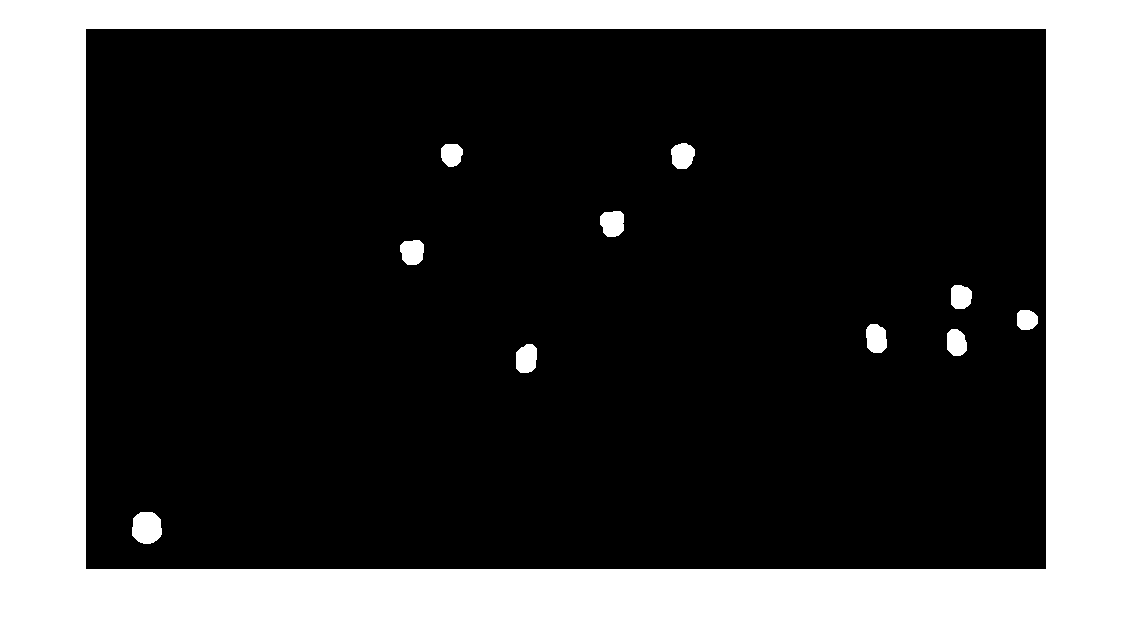}
\caption{(a) Extracting the defending team players (red) (b) Performing morphological operations (fill, open, dilate) to get proper blobs }
\label{fig:red}
\end{figure}

\begin{figure}
\centering
\includegraphics[width=0.45\textwidth]{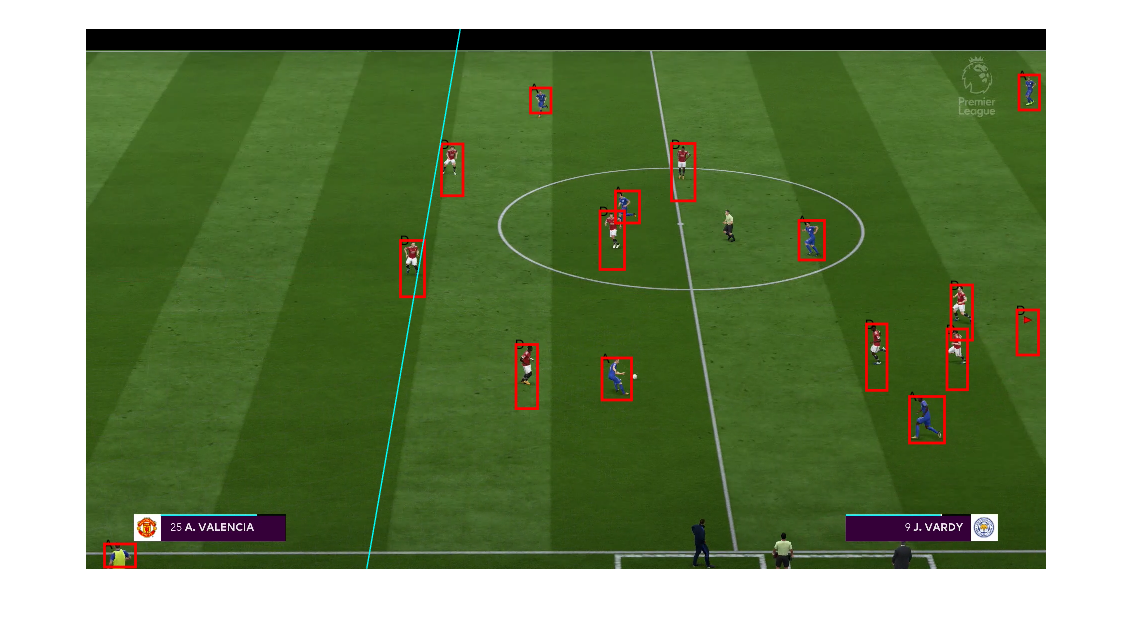}
\caption{Final detections and offside line marking}
\label{fig:final}
\end{figure}

Following our methods above, we now report the stepwise results after each operation.
Figures \ref{fig:hough} through \ref{fig:final} illustrate every step of our method in detail.

Same step is repeated as shown in \ref{fig:red} to obtain the players of the attacking team, but now the colour mask for blue is applied.
Bounding boxes are then drawn on the players as they all form connected components and the offside line is drawn as explained in the previous section. 

\begin{figure}
\centering
\includegraphics[width=0.35\textwidth]{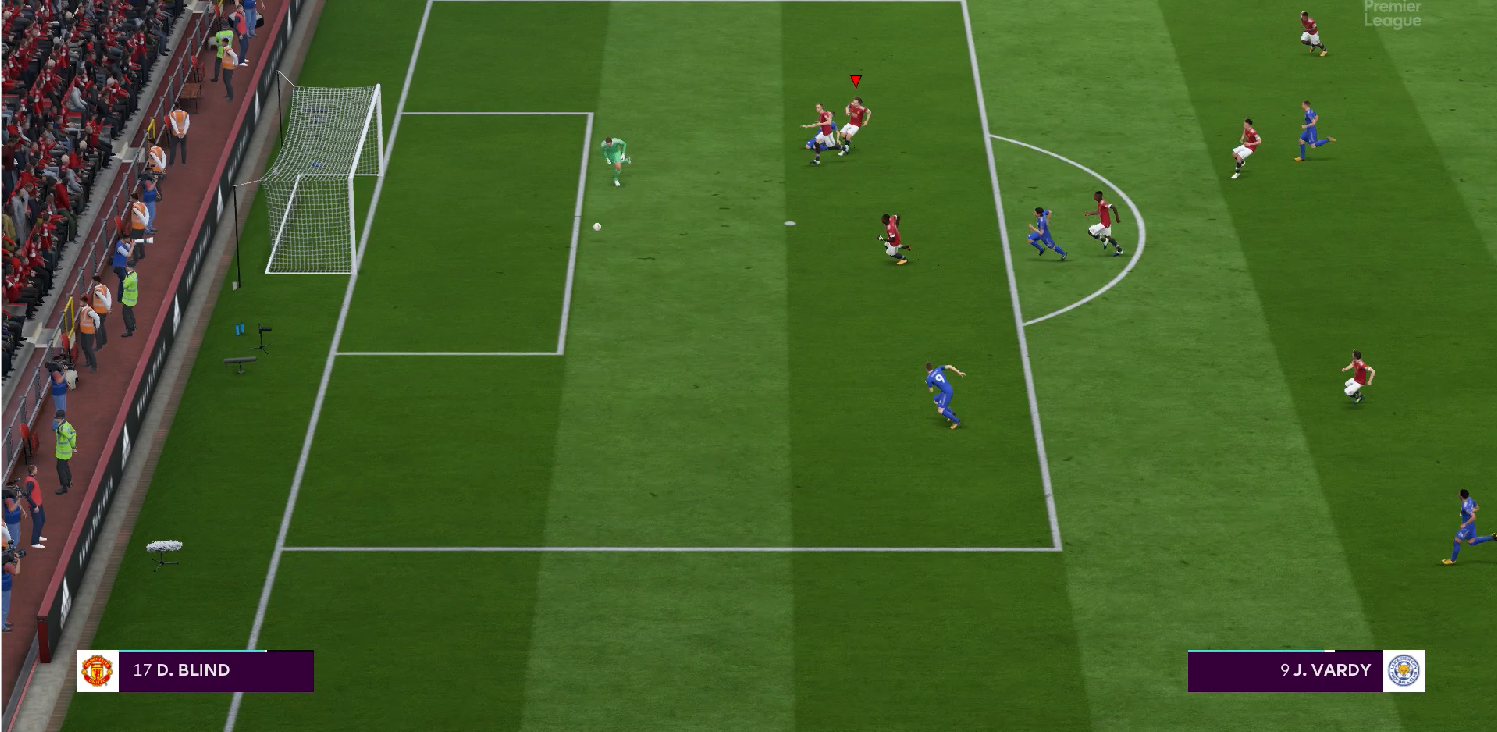}
\includegraphics[width=0.35\textwidth]{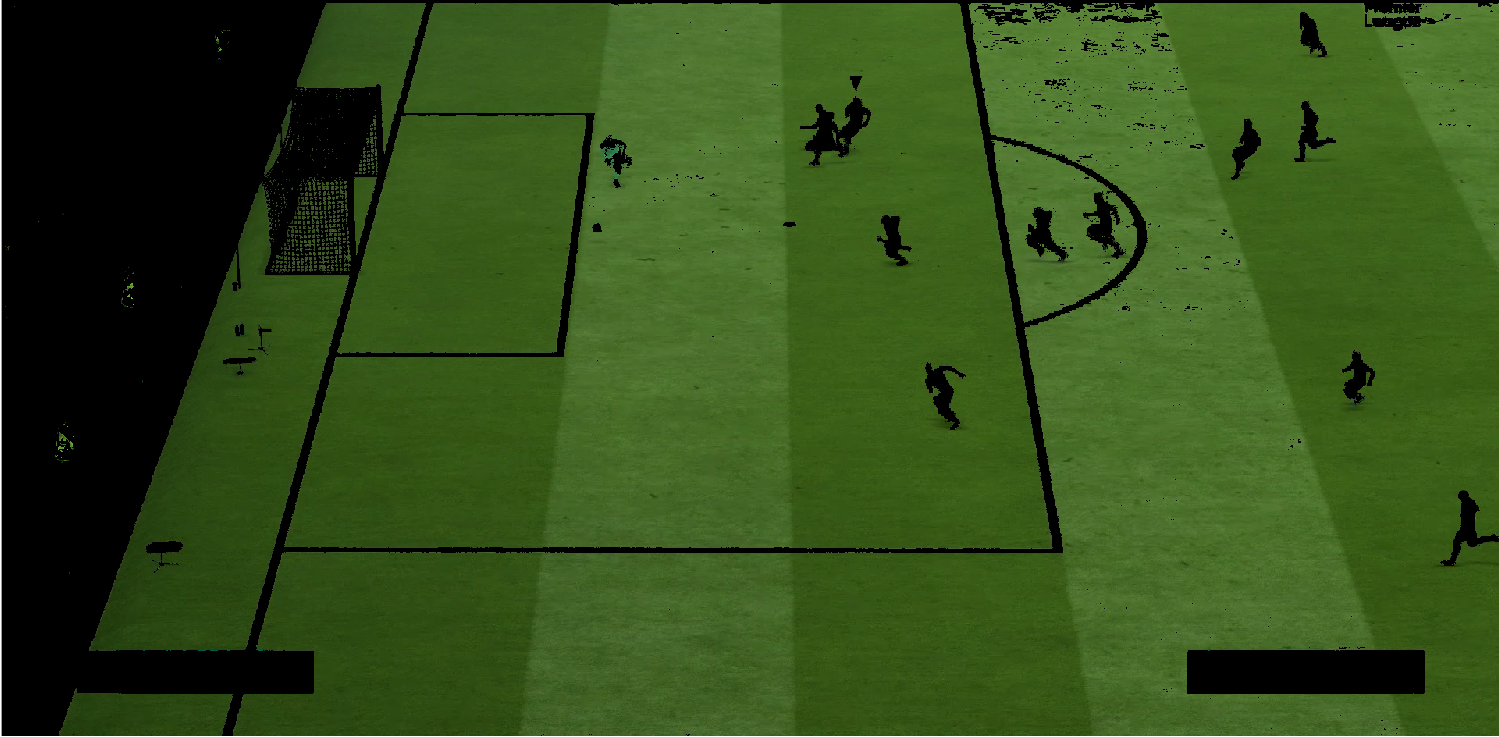}
\caption{Audience stand segmented out}
\label{fig:side}
\end{figure}

Our approach of detecting the active play area also works well with cases where the audience/crowd stands come into the frame as shown in figure \ref{fig:side}.

More details about our implementation and experiments can be found \href{https://github.com/surajkra/Offside_Tracker_EECS504}{here}. Table I also shows some qualitative results. Legend for the table : FP = False Positives, FN = False Negatives, SC = Stadium Configuration, BCC = Boundary Line Connected Components.

\begin{table}[h!]
\caption{Qualitative Comparison}
\begin{tabular}{ |p{2cm}||p{1cm}|p{1cm}|p{1cm}| p{1.5cm}| }
 \hline
 \multicolumn{5}{|c|}{Performance (FP, FN) and Invariance (SC,BCC) Measures} \\
 \hline
\textbf{Architecture} & \textbf{FP} &  \textbf{FN} & \textbf{SC} & \textbf{BCC}\\ \hline
 Proposed   & Low & Low & Invariant & Invariant \\ \hline
 YOLO &   High  & High & Invariant & Invariant \\ \hline
 Morphology only & Low & Medium & Varying & Highly Varying\\
 \hline
 \end{tabular}
 \end{table}

\section{Discussion}
In this section we describe all other algorithms we experimented in the process of identifying the offside lines along with possible future directions. 
\subsection{Why not Deep Learning?}
We first employed the You Only Look Once (YOLO) \cite{c5}network for detecting players in every frame. However the performance was not satisfactory. We speculate the reason to be as follows: YOLO was pre-trained on the MS-COCO dataset and hence isn’t suited for identifying people in our setting where they are highly scaled down. Even though it detects people, this method is not highly robust as the probability of false detections and misses are notably high. But, this model could work significantly well, when it is specifically trained for images from our setting. Due to lack of computational resources devoted for this project and the lack of labeled data specific to our problem, a Deep Learning model couldn’t be trained. However, this qualifies as a promising future work idea. This also extends to other data/compute intensive deep learning methods such as Faster-RCNN and Mask-RCNN.

\subsection{Extending Morphology and Homography}
We also experimented the idea of employing morphological dilation and erosion algorithms on edge detected frames (all background pixels are removed) and setting up a map of all detected points with a top-view 2D field image using homographic transforms. This method was not robust across frames because of the presence of varying boundary lines (white lines) since erosion and dilation operations are subjective to specific frame properties. Also, the mapping from the real coordinates to the 2D top view coordinates is not a straight forward homography problem since it is not a viewpoint change or a direct transformation. Matching SURF features with the reference image didn't work well due to lack of significant interest points in every single frame. An interesting future direction would be to incorporate some ground truth maps/point clouds of the stadium to get accurate top view homographic transformation of the current frame, similar to the NFL first down marker.

\end{document}